\documentclass[journal]{IEEEtran}
\usepackage[utf8]{inputenc}
\usepackage{graphicx}
\usepackage{cite}

\usepackage{amsfonts}
\usepackage{amssymb}
\usepackage{amsmath}
\usepackage{algorithm}
\usepackage{multirow}
\usepackage{multicol}

\usepackage{exscale}
\usepackage{relsize}
\usepackage{algpseudocode}
\usepackage{graphics}
\usepackage{mathrsfs}
\usepackage{siunitx}
\usepackage{url}
\usepackage{lipsum}
\usepackage{array}

\usepackage{caption}

\usepackage{xcolor}
\usepackage{threeparttable}
\usepackage{booktabs}
\usepackage{array}

\DeclareMathOperator*{\argmax}{argmax}
\newcommand{\bm}[1]{\mbox{\boldmath{$#1$}}}

\begin{document}

\title{Neural Network Approximation of Graph Fourier Transforms for Sparse Sampling of Networked Flow Dynamics}

\author{
Alessio~Pagani, Zhuangkun~Wei, Ricardo~Silva, Weisi~Guo
\thanks{A. Pagani is with the Alan Turing Institute, London, UK, e-mail: apagani@turing.ac.uk.}
\thanks{Z. Wei is with the University of Warwick.}
\thanks{R. Silva is with the Alan Turing Institute and University College London.}
\thanks{W. Guo is with Cranfield University and the Alan Turing Institute.}
\thanks{Corresponding author: wguo@turing.ac.uk}
\thanks{Manuscript received XXX; revised XXX.}}



\maketitle
\begin{abstract}

Infrastructure monitoring is critical for safe operations and sustainability. Water distribution networks (WDNs) are large-scale networked critical systems with complex cascade dynamics which are difficult to predict. Ubiquitous monitoring is expensive and a key challenge is to infer the contaminant dynamics from partial sparse monitoring data. Existing approaches use multi-objective optimisation to find the minimum set of essential monitoring points, but lack performance guarantees and a theoretical framework. 

Here, we first develop Graph Fourier Transform (GFT) operators to compress networked contamination spreading dynamics to identify the essential principle data collection points with inference performance guarantees. We then build autoencoder (AE) inspired neural networks (NN) to generalize the GFT sampling process and under-sample further from the initial sampling set, allowing a very small set of data points to largely reconstruct the contamination dynamics over real and artificial WDNs. Various sources of the contamination are tested and we obtain high accuracy reconstruction using around 5-10\% of the sample set. This general approach of compression and under-sampled recovery via neural networks can be applied to a wide range of networked infrastructures to enable digital twins.
\end{abstract}

\begin{IEEEkeywords}
sampling theory, graph Fourier transform, neural networks.
\end{IEEEkeywords}

\IEEEpeerreviewmaketitle

\section{Introduction}\label{sec:Intro}
\IEEEPARstart{C}{ontamination} in drinking water supply arise from natural disasters \cite{GUIKEMA2009855,Pye713}, industrial crime, and terrorism \cite{mays04,doi:10.1080/07900620903392158}. Together, they pose serious risks to the safety and integrity of Water Distribution Networks (WDNs). Contamination spreading in WDNs are governed by both Navier-Stokes dynamics and the topological structure of the network. Heterogeneous elements (e.g. pumps, filters, reservoirs), feedback loops, and the vast size of the network (100,000s of nodes) make the spread process difficult to predict without large-scale simulation and wide-spread data monitoring.

Using sparse sampling data, reconstructing the contaminant dynamics over a vast network could enable timely interventions and help to save lives. The simplest way to detect contaminant would be to install probes in each junction and monitor various dynamic states. However, this is often not possible because of the high installation and maintenance cost and the difficulty in accessing underground pipes retrospectively. This raises the necessity of optimized sensor placement, detecting chemical intrusions and predicting the contamination spread in the shortest time and with the highest accuracy possible.

Optimal sensor placement \cite{Chang12,8306485,6718032} techniques allow operators to track the spread of the contaminants by reconstructing and predicting the spread dynamics. To further reduce the number of sensors, an imperfect reconstruction of the dynamics could be accepted if it guarantees essential KPIs (e.g., low time to detect chemical intrusion, low amount of contaminated water consumed or population affected). This challenge, in first instance, can be framed as a graph signal processing (GSP) \cite{Chen16}, whereby WDNs are flow-based complex networks with a fixed heterogeneous topology (e.g. each node can represent different functions) and multiple coupled dynamic signals (e.g. pressure, contaminant concentration, flow speed). 

\begin{figure*}[t]
     \centering
     \includegraphics[width=1\linewidth]{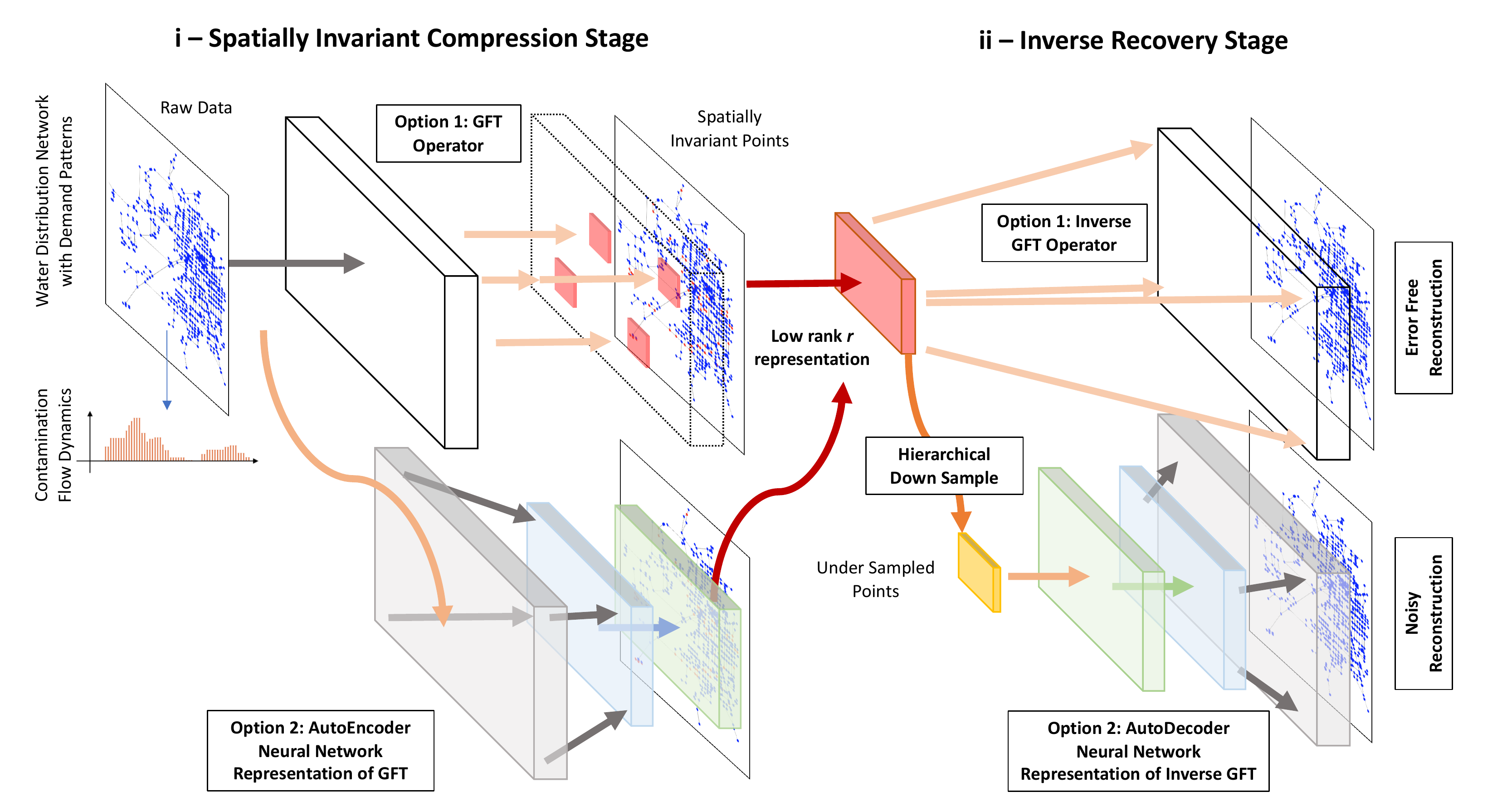}
     \caption{Compressing dynamic contamination data on a water distribution network to spatially invariant monitoring points and recovering the dynamics: (a) full GFT or NN compression and (b) recovery by either inverse-GFT or NN with hierarchical down-sampling option.}
     \label{fig:0}
\end{figure*}

\subsection{State of the Art}
Several studies have been performed on optimizing sensor placement via different perspectives (engineering optimisation, graph-theoretic analysis, and data-driven compression) and to detect contamination sources. We give a brief review of them below.

\subsubsection{Numerical Optimization Approaches}
Rule based multi-objective optimisation considers a number of performance metrics and factors related to both WDN dynamics, as well as accessibility and complexity aspects of the cyber-physical interface \cite{Chang12}. For example, Berry et al. \cite{Berry05} tackled the problem of sensor placement formulation by optimizing the number of sensors that minimize the expected fraction of population at risk. The optimisation approaches include mixed-integer program (MIP), randomized contamination matrix \cite{Kessler98}, and genetic algorithms \cite{Ostfeld04}. Other common approaches revolve around multi-objective optimization frameworks, this gives the capability to reduce the dimensionality of the network through a sensitivity-informed analysis \cite{Fu15} and incorporates uncertainty in the network's demands and Early Winning System operation \cite{SANKARY2017160}. These computational techniques suffer from the lack of explicit relational knowledge between the topological structure and the underlying dynamics with the optimal sampling points. Furthermore, the aforementioned solutions become less feasible for large-scale networks, especially for multiple or diverse contamination dynamics. Whilst computational improvements for multi-objective optimisation in WDNs have been developed \cite{doi:10.1061/WR.1943-5452.0000001, Krause08Opt}, such as a progressive genetic algorithm (PGA), they do not offer performance guarantees nor theoretical insights.

\subsubsection{Graph-Based Analytical Approaches}
Graph Spectral Techniques (GSTs) that identify the most influential points on the base of the topological structure of the networks (e.g. via the Laplacian operator \cite{Archetti15,Giudicianni18,Simone18}) offer theoretical insight between the topological structure of the network and the key monitoring points. These approaches significantly reduce the computation complexity by removing the need of hydraulic simulations \cite{dinardo18, Fu14}, but tend to assume a homogeneous network (e.g. pumps, reservoirs, and junctions are treated equally). The main assumption in topological analysis that doesn't consider the underlying fluid dynamics is that it is assumed that the topology dominates. As such, it is important to create an approach that considers both the complex network topology and the contamination signals. The challenge with WDNs is that the underlying Navier-Stokes dynamics with dynamic Reynolds numbers is high dimensional and highly non-linear \cite{Guymer16}. As such, an analysis of the optimal sampling points as a function of both the network topology and the dynamic equations is non-trivial. 

\subsubsection{Data-Driven Approaches}
One approach that considers the data-structure is compressed sensing (CS) \cite{Du15CS, McCann15, Xie17CS}, which compresses the data by transforming them into a sparse domain. However, the main challenge lies in the unknown of positions of such sparse non-zeros elements in the transformed data, which will inevitably lead to an approximately $(N+K-r)\times r/K$ nodes for monitoring (for a data matrix $\mathbf{X}$ with $N$ nodes, $K$ time-step and $r=rank(\mathbf{X})$) \cite{5730578, 8839864}. This is not to mention that most of the CS approaches do not guarantee the unchanged nodes for sensor deployment. 

To further reduce the number of sampling nodes, our previous work in \cite{8839864} proposed a data-driven GFT sampling method, which is able to characterize the data matrix into an $r$-bandlimited space, and thereby ensures the recovery accuracy with the monitoring of only $r$ orthogonal nodes. However, the overlooks of the unknown and latent transitions among states limit the further reduction of the number of monitoring nodes. This raises the necessity of the understanding of such relations underlying the data, and we do so by the run of a Neural Network. 

\subsubsection{Graph Neural Networks}
Graph neural networks (GNNs) were initially proposed by Gori et al. \cite{gori1555942} and Scarselli et al. \cite{scarselli4700287}. These early studies fall into the category of recurrent graph neural networks (RecGNNs). Convolutional neural networks (ConvNNs) were then introduced by Bruna et al. \cite{BrunaZSL13} and gained popularity \cite{henaff2015deep,levie8521593}. In the last couple of years, several works focused on using machine learning techniques with arbitrarily structured graph data \cite{Kipf16,defferrard2016}.

A comprehensive survey of NN applied to graphs can be found in \cite{wu2019comprehensive}, including the most recent techniques based on Graph Autoencoders (GAEs): GAEs map nodes into a latent feature space and decode graph information from latent representations. They can be used to learn network embeddings or generate new graphs.

In our work, we improve the state of the art exploiting a GAE-based technique to optimally sample the network considering the graph- and dynamics- domain, with the aim to reconstruct the dynamics in all the nodes of the graph.
NN alone, in fact, cannot inform us which nodes are optimal for sampling, nor how many nodes are needed. This is why we are using the GFT to inform and drive the NN, which is novel. 

\paragraph{Neural Networks for WDN applications}
Neural Networks (NN) are increasingly being used in WDN applications, especially for water supply issues and predictions relative to chemical disinfectants and contaminants. A synopsis of NN methods, including the design and operation of WDNs, is provided in review article \cite{Mosetlhe15}. Proposals for a neural network in the assessment of pressure losses in water pipes are proposed in \cite{Czapczuk17, Dawidowicz18}. Cuesta Cordoba et al. \cite{CORDOBA2014399} used an NN to predict chlorine decay using historical data. Similarly, Andrade et al. \cite{Andrade13}, estimated the disinfectant concentration at the relevant nodes using NN with the aim of improving WDNs design. These approaches, however, either work well only on a pre-selected subset of nodes considered important, or require a high number of sensors to work on the whole water infrastructure. Moreover, they are not designed for working in real time, excluding the capability to promptly predict chemical intrusion and spreading.

\begin{figure*}[!ht]
    \centering
    \begin{minipage}[b]{\columnwidth}
      \centering
      \includegraphics[width=\textwidth]{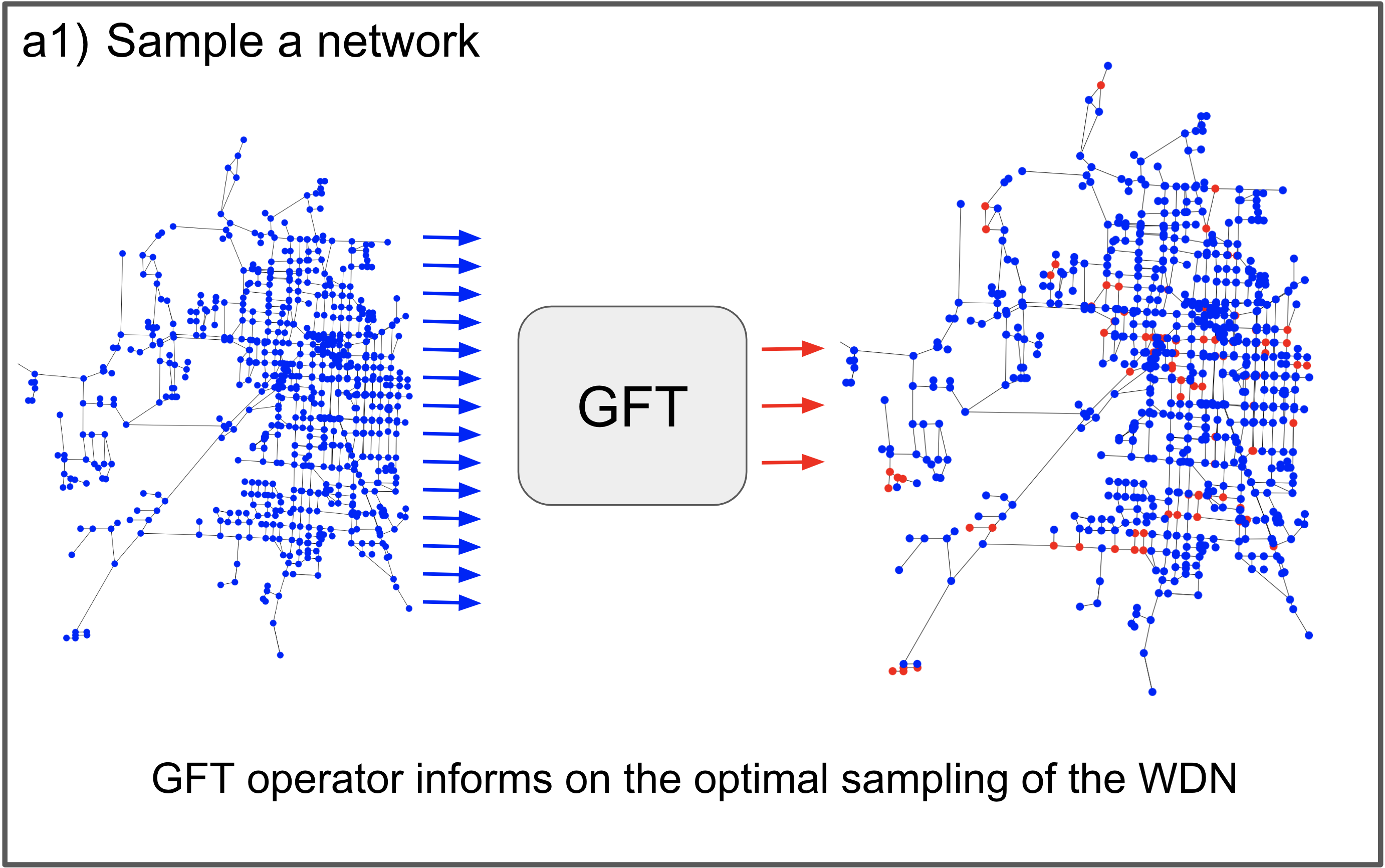}
    \end{minipage}
    \begin{minipage}[b]{\columnwidth}
      \centering
      \includegraphics[width=\textwidth]{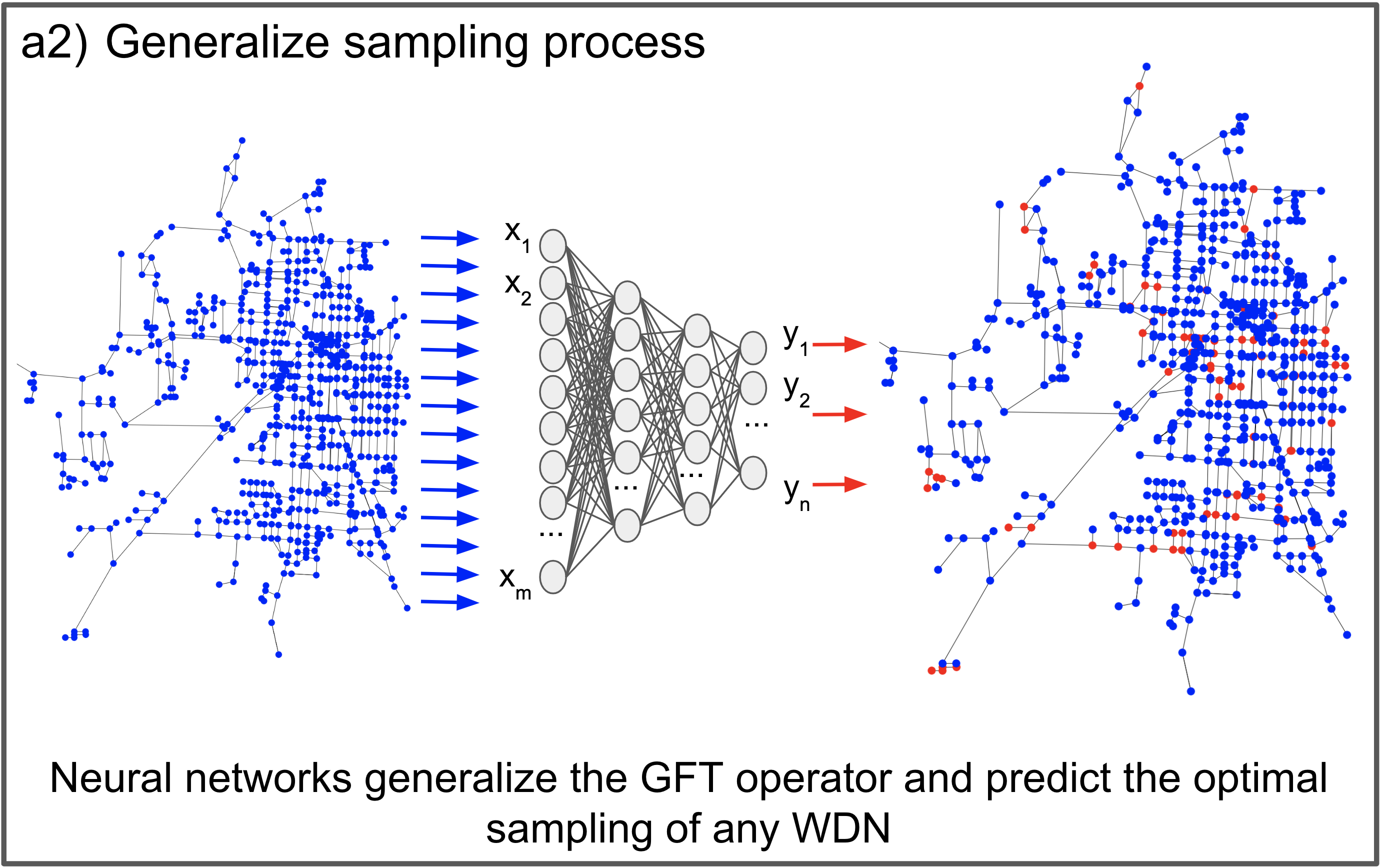}
    \end{minipage}
    \begin{minipage}[b]{\columnwidth}
      \centering
      \includegraphics[width=\textwidth]{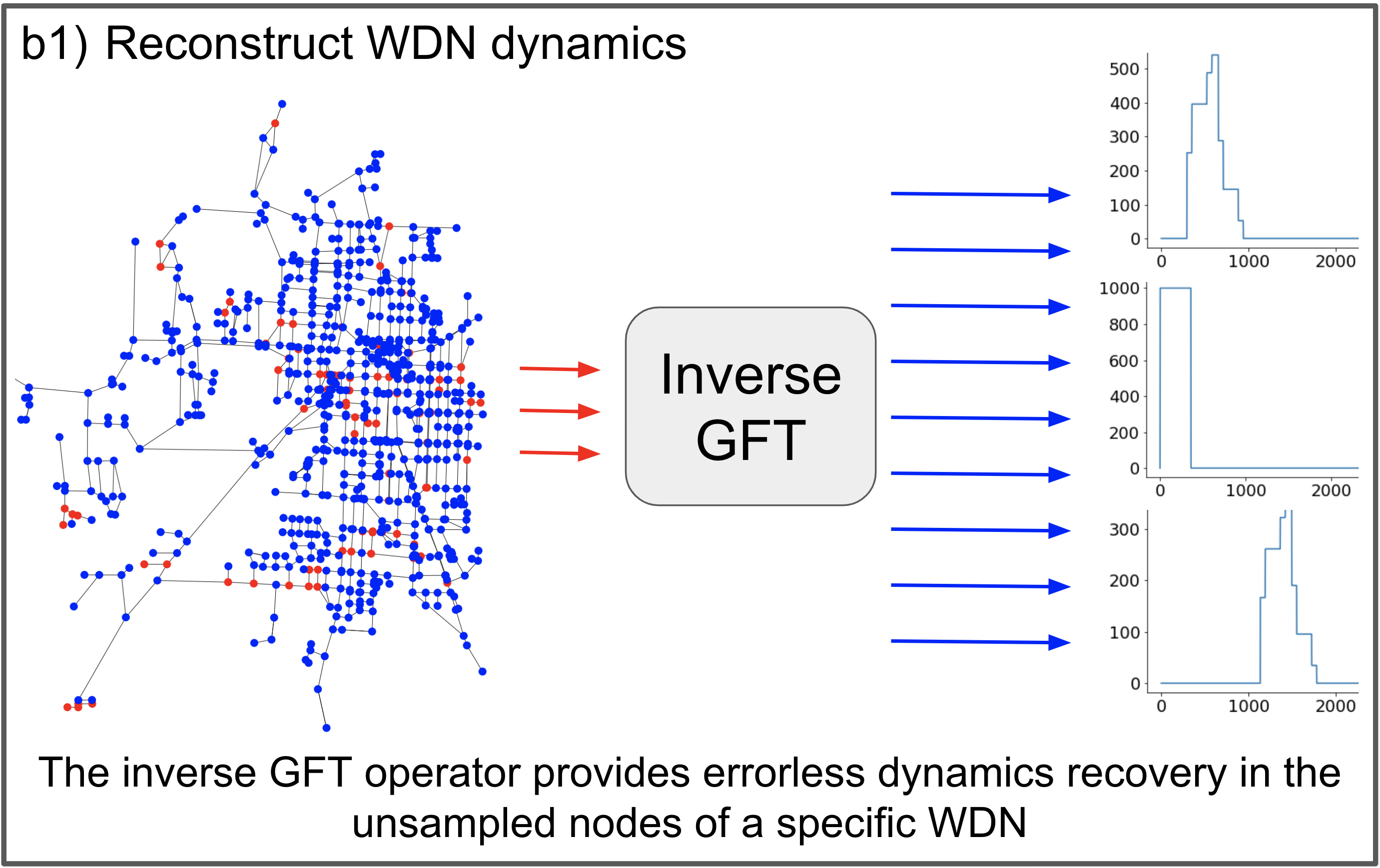}
    \end{minipage}
    \begin{minipage}[b]{\columnwidth}
      \centering
      \includegraphics[width=\textwidth]{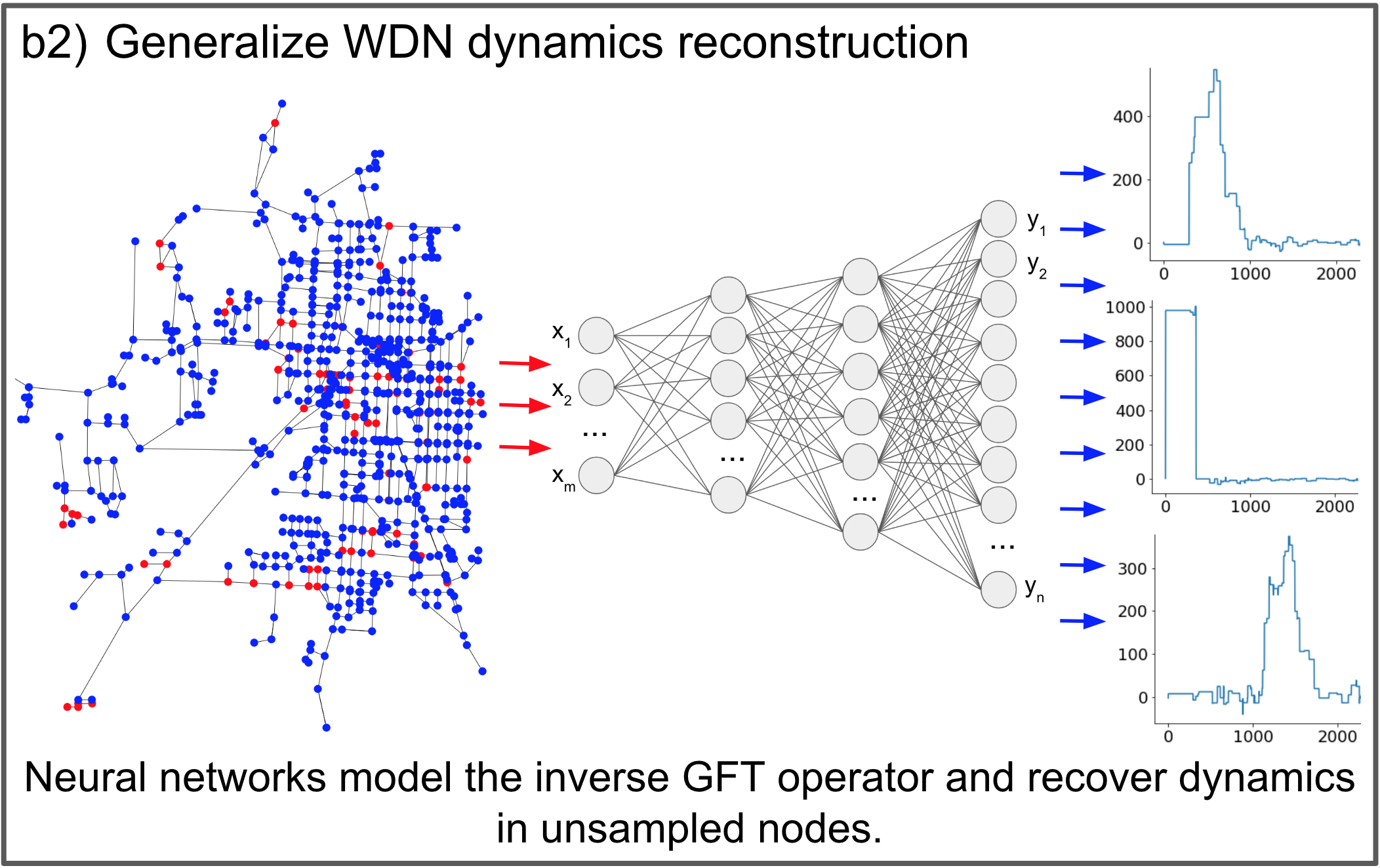}
    \end{minipage}
    \caption{Sampling process. The GFT informs on the optimal number of nodes required to reconstruct the dynamics without errors (a1).
    A Neural Network (NN) is used to generalize the process and reduce the number of sampled nodes (a2).
     The inverse GFT operator is used to reconstruct the dynamics in all the nodes of a specific network (b1). NN can be used to model the inverse GFT operator and reconstruct the dynamics for every given input (b2).
     Using NN to sub-sample the GFT ensemble or reconstruct unseen signals can introduce errors in the dynamics reconstruction, with a trade-off between size of the final ensemble and accuracy.
    }
    \label{fig:nn_architecture}
\end{figure*}

\subsection{Contribution}
We aim to use sparse data to recover the contaminant spreading dynamics in the WDNs. In this work, we first use GFT to derive the operator that reveals the minimum number of nodes needed to sample the WDN and recover its dynamics with minimum error (see Fig. \ref{fig:0}). This approach was outlined in our recent preprint \cite{Wei19} (Fig. \ref{fig:nn_architecture}(a)), which minimized sensor deployment number at the penalty of relying on the inverse GFT operator to recover the dynamics, only with the full set of sampled data. Here we expand on this by under-sampling the GFT derived set of minimum nodes to create noisy measurements (Fig. \ref{fig:nn_architecture}(b)). By developing a neural network (NN), we can sample from an extremely sparse set that can recover the essential noisy contamination dynamic trend. This is useful because often we are interested in whether a contamination has exceeded a threshold, but not necessarily its full dynamic response.

There is a natural trade-off between reconstruction accuracy and size of the sampled nodes ensemble used for the training, but we show that NNs can significantly reduce the number of required nodes (compared to the pure GFT approach) maintaining a high accuracy. The final nodes ensemble is the the minimal set of nodes that has to be equipped with probes and sensors for monitoring. We reconstruct the chemical spread dynamics using two scenarios: known potential contamination source (e.g., factories or plants, maintenance works, known dangerous areas) and unknown contamination source (e.g., terrorist attacks).

The final outcome of this work is a general framework able to optimally sample a network with complex flow dynamics (e.g. PDEs with dynamic parameters and feedback loops) and then reconstruct the flow dynamics in all the network nodes using only the sparse partial dynamics at optimal points. The framework is composed of two main parts: (1) an encoder (via GFT or NN) - that compress the network considering its topology and dynamics, and a decoder (via inverse-GFT or NN) - that reconstructs the flow dynamics.

The rest of paper is structured as follows. In Section II, we describe the initial sampling process based on GFT and the subsequent sampling reductions that is used as input for the NN models. We then introduce the NN architecture and the WDN simulations. In Section III, we show the results of the dynamics reconstructions using the NN models. In Section IV, we conclude the paper and discuss the potential future areas of research.

\section{Methods}

We first use GFT to develop an operator that can transform the WDN contamination spread dynamic into a band-limited set of sampling nodes for guaranteed inference performance. We then use the sampling set to hierarchically down-sample and recover the noisy dynamics via a neural network.

\subsection{GFT Sampling Process}
\label{sec:sampling_process}
Graph sampling theory over complex network \footnote{a complex network is a graph (network) with non-trivial topological features} aim at sampling and recovering the time-varying networked signals, denoted as $\mathbf{X}\in\mathbb{R}^{N\times K}$ that are $(r<N)$-bandlimited to a given GFT operator, denoted as $\mathbf{F}^{-1}$. Here, we consider a \textit{static topology} with \textit{time-varying dynamic signals}. The $N$ rows of $\mathbf{X}$ present the time-varying signals on $N$ nodes, and $K>N$ denotes the total number of time-indices. We say $\mathbf{X}$ is $r$-bandlimited to $\mathbf{F}^{-1}$, if and only if its GFT response $\tilde{\mathbf{X}}=\mathbf{F}^{-1}\cdot\mathbf{X}$ has only $r$ nonzero rows. Denote $\mathcal{R}$ is the set of subscripts of the nonzero rows in $\tilde{\mathbf{X}}$, and $\mathcal{V}=\{1,\cdots,N\}$. Then, we say there exists a subset $\mathcal{S}\subset\mathcal{V}$ such that:
\begin{equation}
\label{gft_th1}
    \mathbf{X}=\mathbf{F}_{\mathcal{V}\mathcal{R}}\cdot (\mathbf{F}_{\mathcal{S}\mathcal{R}}^T\cdot\mathbf{F}_{\mathcal{S}\mathcal{R}})^{-1}\cdot\mathbf{F}_{\mathcal{S}\mathcal{R}}^T\cdot\mathbf{X}_{\mathcal{S}\mathcal{K}},
\end{equation}
if and only if:
\begin{equation}
\label{gft_th2}
    rank(\mathbf{F}_{\mathcal{S}\mathcal{R}})=|\mathcal{R}|=r. 
\end{equation}
In Eqs. (\ref{gft_th1})-(\ref{gft_th2}), $\mathbf{X}_{\mathcal{S}\mathcal{K}}$ denotes the sample of $\mathbf{X}$ from nodes that belongs to $\mathcal{S}$. $\mathbf{F}_{\mathcal{S}\mathcal{R}}$ denotes the selection of the matrix $\mathbf{F}$ with row indices from set $\mathcal{S}$, and column indices from set $\mathcal{R}$. The sampling and recovering processes can be pursued after (I) designing the GFT operator, and (II) the selection of $\mathcal{S}$ satisfying Eq. (\ref{gft_th2}).

\subsubsection{GFT Operator Design}
The design of the GFT operator $\mathbf{F}^{-1}$ is borrowed from the QR factorization. To be specific, as we derive the maximally linearly independent columns of $\mathbf{X}$, denoted as $\mathbf{X}_{\mathcal{V}\mathcal{M}}=[\mathbf{x}_{m_1},\cdots,\mathbf{x}_{m_r}]$, the GFT operator can be computed as:
\begin{equation}
\label{qr}
    \mathbf{F}^{-1}=\mathbf{Q}^{-1},
\end{equation}
where $\mathbf{X}_{\mathcal{V}\mathcal{M}}=\mathbf{Q}\cdot\mathbf{R}$. This $\mathbf{F}^{-1}$ ensures the $r$-bandlimited property of $\mathbf{X}$, since:
\begin{equation}
\begin{aligned}
    \tilde{\mathbf{X}}=&\mathbf{F}^{-1}\cdot\mathbf{X},\\
    \overset{\text{(a)}}{=}&\mathbf{F}^{-1}\cdot\left[\mathbf{X}_{\mathcal{V}\mathcal{M}},~\mathbf{X}_{\mathcal{V}\mathcal{M}}\cdot\bm{\Pi}\right],\\
    \overset{\text{(b)}}{=}&\left[\mathbf{R},~\mathbf{R}\cdot\bm{\Pi}\right].
    \label{proof bandlimited}
\end{aligned}
\end{equation}
In Eq. (\ref{proof bandlimited}), (a) holds for that each column of $\mathbf{X}$ can be expressed by the columns from $\mathbf{X}_{\mathcal{V}\mathcal{M}}$ multiplied with an $r\times(K-r)$ matrix $\bm{\Pi}$, since $rank(\mathbf{X}_{\mathcal{V}\mathcal{M}})=rank(\mathbf{X})=r$. (b) indicates that only the first $r$ rows of $\tilde{\mathbf{X}}$ are non-zero, as $\mathbf{R}$ is the upper triangular matrix with $rank(\mathbf{R})=r$.

\subsubsection{Signal Recovery}
After the computation of the GFT operator in Eq. (\ref{qr}), one needs to ensure a complete recovery is to select $\mathcal{S}$ that satisfies Eq. (\ref{gft_th2}). One can refer to \cite{chen2015discrete,p2008, Sandryhaila14, anis2014towards, Chen15, wang2015generalized, anis2016efficient, Chen16, wang2018optimal, chamon2018greedy, ortega2018graph} for details. Here, in order to achieve a robust sampling scheme on nodes, we consider the selection of $\mathcal{S}$ that maximizes the minimum singular of $\mathbf{F}_{\mathcal{S}\mathcal{R}}$, i.e., 
\begin{equation}
    \mathcal{S}_{\text{opt}}=\argmax_{\mathcal{S}\subset\mathcal{V}}\sigma_{\text{min}}\left(\mathbf{F}_{\mathcal{S}\mathcal{R}}\right),
    \label{s_option}
\end{equation}
where $\sigma_{\text{min}}(\cdot)$ denotes the smallest singular value. As such, the importance of the nodes in $\mathcal{S}$ can be ranked with the descending order of the singulars.

\subsection{Sampling Reduction Using Neural Networks}
As discussed in the previous section, the initial sampling is conducted using the GFT analysis that exploits the low-rank property to optimally sample junction nodes in WDNs. Using GFT, it is possible to fully recover network dynamics of a specific injection scenario (see Section \ref{sec:injection_scenarios} for details on the scenarios) using a subset of data sampled at the identified nodes. The identified nodes are used as initial subset, hereafter called the \textbf{GFT dataset}. The GFT dataset is different for each possible source of contamination (injection location).

In order to generalize this approach, we use a neural network (NN) to model the GFT and detect the optimal nodes to sample. The sampling reduction process depends on the prior knowledge of the contaminant source: (I) when the source is known, a \textit{injection-specific approach} can be used, otherwise (II) a \textit{general approach} is required. The new reduced dataset, hereafter called \textbf{sampling dataset}, is then used to train a second NN (represent the inverse-GFT) for reconstructing the dynamics.

\subsubsection{Injection-Specific Approach}
When the chemical source is known, the relative GFT dataset can be used as initial subset. In a GFT dataset the nodes are ranked in order of importance for the reconstruction of the signal. For this reason, the \textbf{injection-specific sampling datasets} are generated removing from the GFT dataset, one by one, the nodes with the lower rank. Each newly created sampling dataset is used to train a NN and the model performance is evaluated. The reduction process is repeated until the reconstruction accuracy drops below a given threshold.

\subsubsection{General Approach}
When unknown, we use two techniques to define the sampling dataset, one based on the nodes frequency in the GFT datasets, the other based on their importance (rank) in each GFT dataset.

\paragraph{GFT Frequent Nodes Dataset}
this approach revolves around the selection of the more frequent nodes in the GFT datasets of the different injection scenarios.
The \textbf{GFT frequent nodes dataset (GFT-F)} is created by counting the times each node appears in the different GFT datasets and selecting only the nodes which appear more than a given threshold. Different threshold are discussed in the results.

\paragraph{GFT Important Nodes Dataset}
this approach concerns the selection of the most important nodes in each injection scenario. In order to create a \textbf{GFT important nodes dataset (GFT-I)}, we consider only the $n$ most important nodes of each GFT dataset. For example, for $n=1$, we select the node with the highest ranking in each GFT dataset (i.e., each injection scenario). This could lead to big sampling datasets even with a small $n$, however the most important nodes are usually shared among different GFT datasets, for this reason the final GFT-I datasets are significantly smaller than the number of different injection scenarios.

\subsection{GFT Datasets Selection}
The number of GFT datasets increases linearly with the number of injection points in the network. The two techniques proposed in the previous section are used to reduce the number of nodes to be monitored, however they do not reduce the number of experiments required (see injection scenarios in Sec. \ref{sec:injection_scenarios} for more details) and the size of the subsequent dataset used for training the neural networks. As the network increases this process can rapidly become computationally very intensive. For this reason, some optimization techniques can be applied to the training dataset. In this paper, we filtered the GFT datasets removing those that are subsets of other GFT datasets. This approach does not change the nodes selection process (i.e., the GFT-F and GFT-I datasets) and the accuracy, however it speeds up the neural network training process.

\subsection{Deep Neural Network Architectures}
\label{sec:neural_network_architecture}

\subsubsection{Encoder for Network Optimal Sampling}
The GFT ensemble created using the GFT operator is specific for a WDN, to overcame this limitation we generalize the problem using a neural network (NN) classifier to model the GFT operator and optimally sample networks with dynamic flows.
This process is similar to encoders used in autoencoders (AE) \cite{Kramer91}, with the difference that our final compression has to be related with physical nodes in the network. To deal with this constraint, although this NN behaves as an encoder, the output layer has the same size of the input layer. The purpose of the NN is to classify the network nodes: each node is associated with a neuron in the input and output layer. While the input layer is fed with node dynamics time series ($\mathbf{X}_i$), the output layer classifies the nodes as important (1) or not important (0) ($\mathbf{y}_i$).
\noindent
For a given NN the input layer is therefore defined as:
$$ \mathbf{X} = [\mathbf{X}_1, ..., \mathbf{X}_n]$$
\noindent
while the output layer is defined as:
$$ \mathbf{Y} = [\mathbf{y}_1, ..., \mathbf{y}_n]$$
\noindent
where $n$ is the number of junctions in network. The NN are trained and tested using GFT optimal ensembles. The optimal number of layers and neurons has been extensively analyzed  and identified experimentally. 


\subsubsection{Decoder for Dynamics Reconstruction}
We trained a NN for the reconstruction of the chemical spread dynamics for each sampling dataset generated. The NNs are feed-forward deep neural networks, with multiple hidden layers and an increasing number of neurons. The optimal number of layers and neurons has been extensively analyzed and identified experimentally. It is worth mentioning that the topology of the WDN is implicit in the training dataset, hence learned by the NNs without need of NN architectures specifically designed for graph learning. Those architectures, however, may be required for other related applications (e.g., leakage detection).

A sensor is installed in each node that belongs to the sampling dataset ($\mathbf{x}_i$), then the NNs are fed with the sensor readings over time. Hence, the input layer is a set of neurons, one for each probe installed in the WDN. The output layer is the estimated concentration of chemical in all the junctions ($\mathbf{y}_i$) of the WDN.

For a given NN the input layer is therefore defined as:
$$ \mathbf{X} = [\mathbf{x}_1, ..., \mathbf{x}_m]$$
\noindent
where $m$ is the size of the sampling subset. While the output layer is defined as:
$$ \mathbf{Y} = [\mathbf{y}_1, ..., \mathbf{y}_n]$$
\noindent
where $n$ is the number of junctions in network.

\subsection{Chemical Injection Simulations}
\label{section:chemical_injection_sim}
The simulations are executed using WNTR (Water Network Tool for Resilience) \cite{WNTR}. WNTR is an EPANET \cite{EPANET2} compatible Python package designed to simulate and analyse resilience of WDNs, it performs extended-period simulation of hydraulic and water-quality behaviour within pressurized pipe networks. This package also supports the simulation of spatially and temporally varying water demand, constant or variable speed pumps, and the minor head losses for bends and fittings. The modelling provides information such as flows in pipes, pressures at junctions, propagation of a contaminant, chlorine concentration, water age, and even alternative scenario analysis.

\subsubsection{Injection Scenarios}
\label{sec:injection_scenarios}
in this work, an \textbf{injection scenario} is a WNTR simulation where a chemical is injected, for a predefined amount of time, in a specific junction of the network. The chemical spreads in the network following the water dynamics and the water demand and it is finally expelled by demand junctions. The simulation ends when the chemical is fully expelled from the network.


\section{Results and Discussion}

In this section we discuss the results obtained using the two approaches to reconstruct the flows dynamics presented in the previous sections: injection-specific sampling, and general sampling. The first can be used for optimal sampling when the source of the injection is known, the second is used when the source is unknown.

The results are compared with two sampling baselines: random sampling and Laplacian sampling. In the \textbf{random sampling} a predefined percentage of WDN nodes is randomly selected. In the \textbf{Laplacian sampling}, the WDN nodes are selected according to their Laplacian rank.

\subsection{Dynamics Reconstruction}
\begin{figure}[!ht]
    \centering
    \begin{minipage}[b]{0.49\columnwidth}
      \centering
      \includegraphics[width=\textwidth, trim={0 0 6.5cm 5cm},clip]{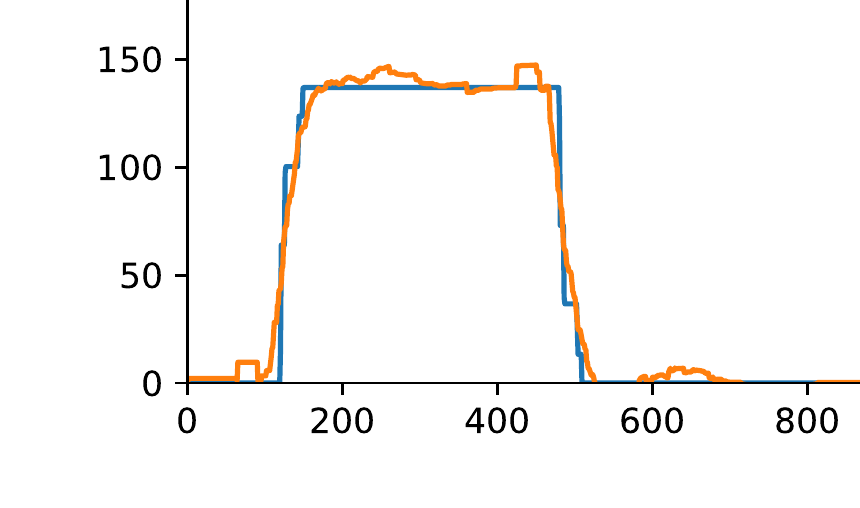}
      GFT dataset.
    \end{minipage}
    \hfill
    \begin{minipage}[b]{0.49\columnwidth}
      \centering
      \includegraphics[width=\textwidth, trim={0 0 6.5cm 5cm},clip]{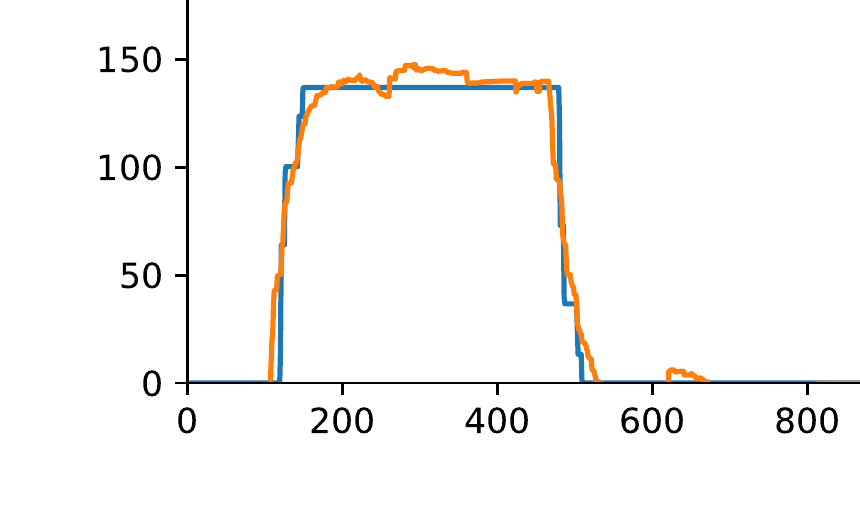}
      75\% GFT dataset.
    \end{minipage}
    \hfill
    \begin{minipage}[b]{0.49\columnwidth}
      \centering
      \vspace{0.8cm}
      \includegraphics[width=\textwidth, trim={0 0 6.5cm 5cm},clip]{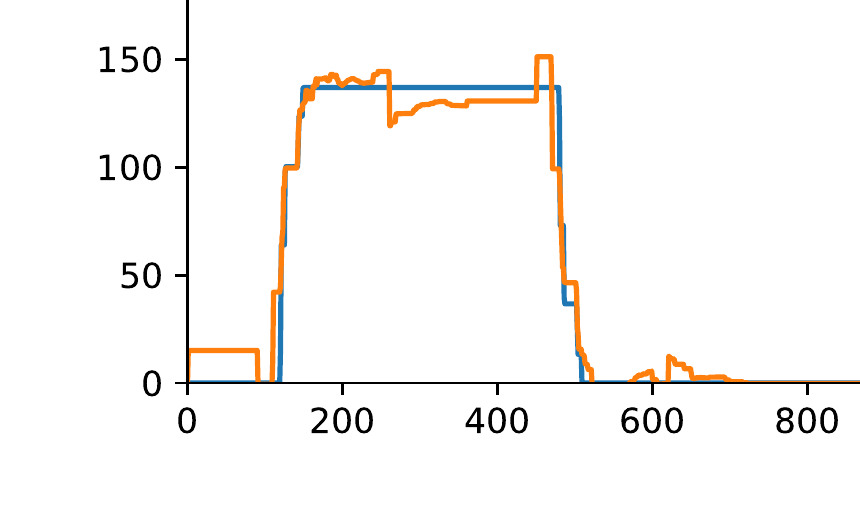}
      50\% GFT dataset.
    \end{minipage}
    \hfill
    \begin{minipage}[b]{0.49\columnwidth}
      \centering
      \vspace{0.8cm}
      \includegraphics[width=\textwidth, trim={0 0 6.5cm 5cm},clip]{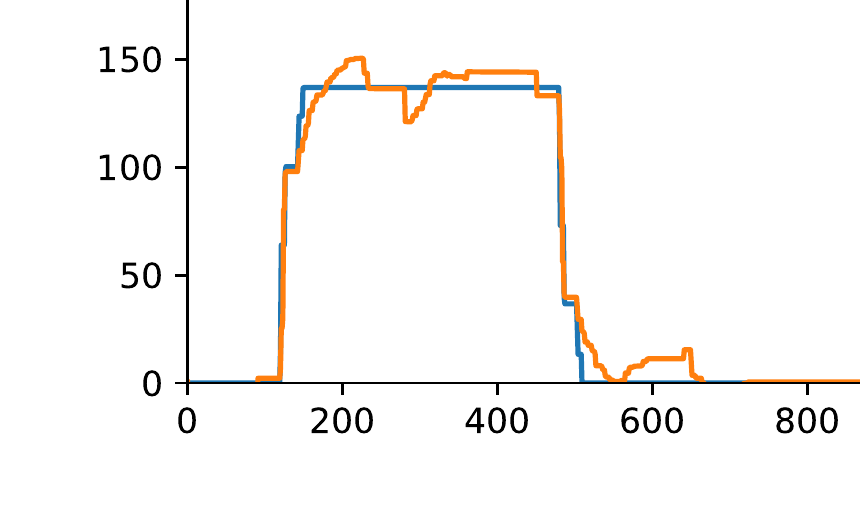}
      25\% GFT dataset.
    \end{minipage}
    \hfill
    \caption{Example 1 - Reconstruction of the dynamics using different portions of the GFT dataset. Original signal in blue, reconstructed signal in orange.}
    \label{fig:example1}
\end{figure}

\begin{figure}[!ht]
    \centering
    \begin{minipage}[b]{0.49\columnwidth}
      \includegraphics[width=\textwidth, trim={0 0 6.5cm 5cm},clip]{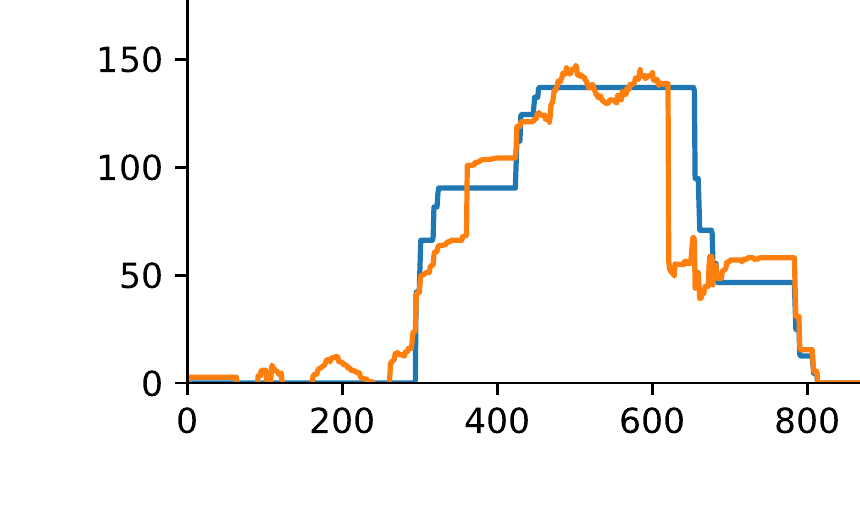}
      \centering
      GFT dataset.
    \end{minipage}
    \hfill
    \begin{minipage}[b]{0.49\columnwidth}
      \centering
      \includegraphics[width=\textwidth, trim={0 0 6.5cm 5cm},clip]{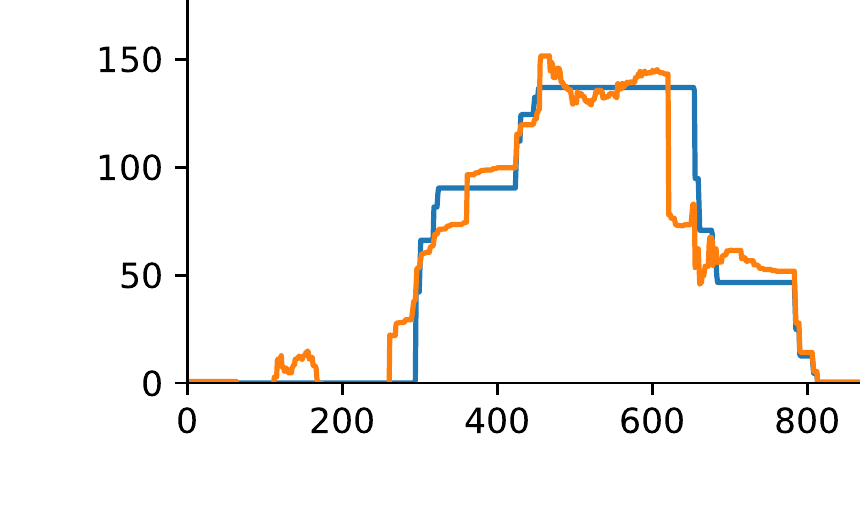}
      75\% GFT dataset.
    \end{minipage}
    \hfill
    \begin{minipage}[b]{0.49\columnwidth}
      \centering
      \vspace{0.8cm}
      \includegraphics[width=\textwidth, trim={0 0 6.5cm 5cm},clip]{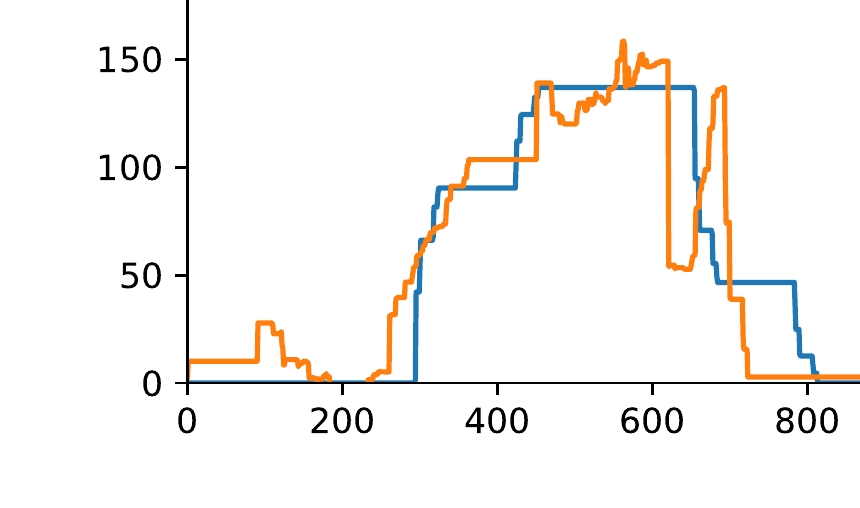}
      50\% GFT dataset.
    \end{minipage}
    \hfill
    \begin{minipage}[b]{0.49\columnwidth}
      \centering
      \vspace{0.8cm}
      \includegraphics[width=\textwidth, trim={0 0 6.5cm 5cm},clip]{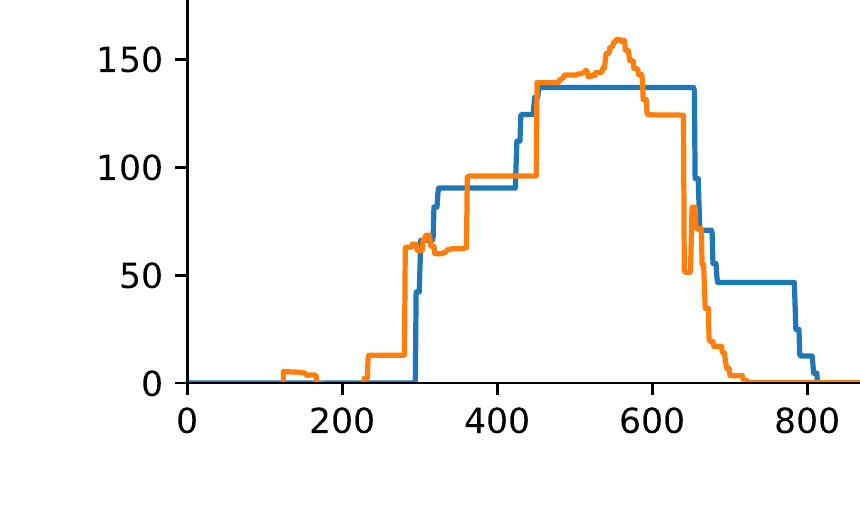}
      25\% GFT dataset.
    \end{minipage}
    \hfill
    \caption{Example 2 - Reconstruction of the dynamics using different portions of the GFT dataset. Original signal in blue, reconstructed signal in orange.}
    \label{fig:example2}
\end{figure}

When the source is known, the GFT dataset can be hugely reduced without significant loss of accuracy. In fact, the NN is able to reconstruct the signal using a few dynamics in specific nodes. Accurate reconstructions are obtained, on average, using 20\% of the GFT dataset (i.e., around 5-10\% of all the WDN nodes has to be monitored, depending on the injection point). Two examples of signals reconstructed using NN models are shown in Fig. \ref{fig:example1} and Fig. \ref{fig:example2}: the concentration over time of a chemical component in two (not-monitored) junctions are reconstructed (in blue the original signal, in orange the reconstructed one).

When the chemical source is unknown, or a general platform able to reconstruct the dynamics in the whole network is required, the injection specific approach cannot be used, instead one of the general approaches is required. While a source specific approach requires, on average, 5-10\% of the WDN nodes to reconstruct the dynamics, the general approaches require several more nodes: among the tested sampling techniques, the best one is the selection of the most frequent nodes in the GFT datasets (GFT-F approach): given a GFT-F dataset, the NN requires, on average, to monitor 50-55\% of the WDN nodes (junctions) for near error-less dynamic reconstruction. The same accuracy is reached using the GFT-I dataset and monitoring 70-75\% of the WDN nodes. More details on the percentage of dynamics correctly reconstituted for a given percentage of WDN nodes is provided in Section \ref{sec:sensitivity_and_specificity}. Both the approaches performed better than the baselines: the signal reconstruction requires respectively 75-80\% of the WDN nodes using the Laplacian ranking and 80-90\% of the WDN nodes using random sampling.

\begin{figure}[!ht]
    \centering
    \begin{minipage}[b]{\columnwidth}
      \includegraphics[width=\textwidth]{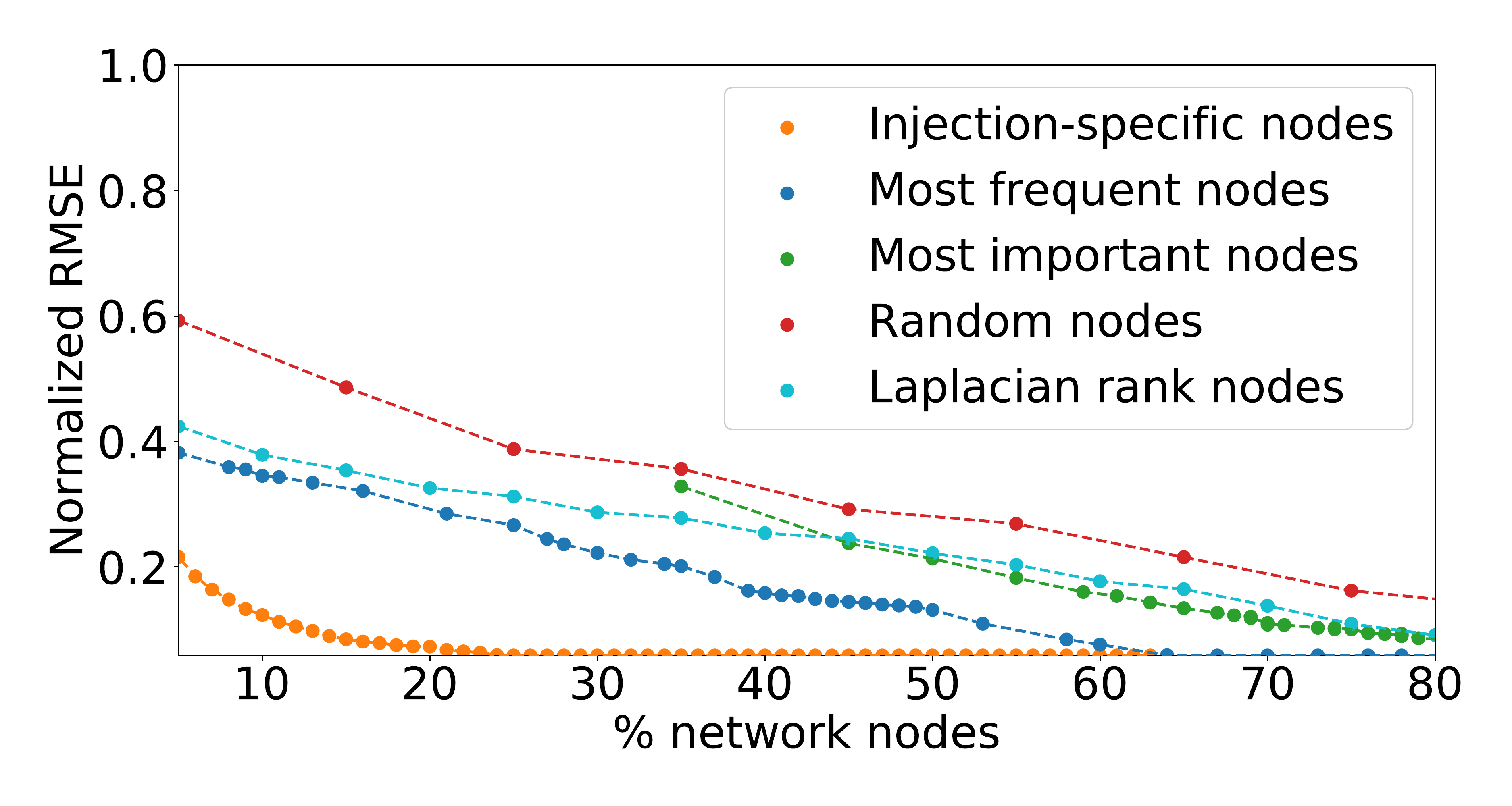}
      \centering
      Average RMSE.
    \end{minipage}
    \hfill
    \begin{minipage}[b]{\columnwidth}
      \centering
      \includegraphics[width=\textwidth]{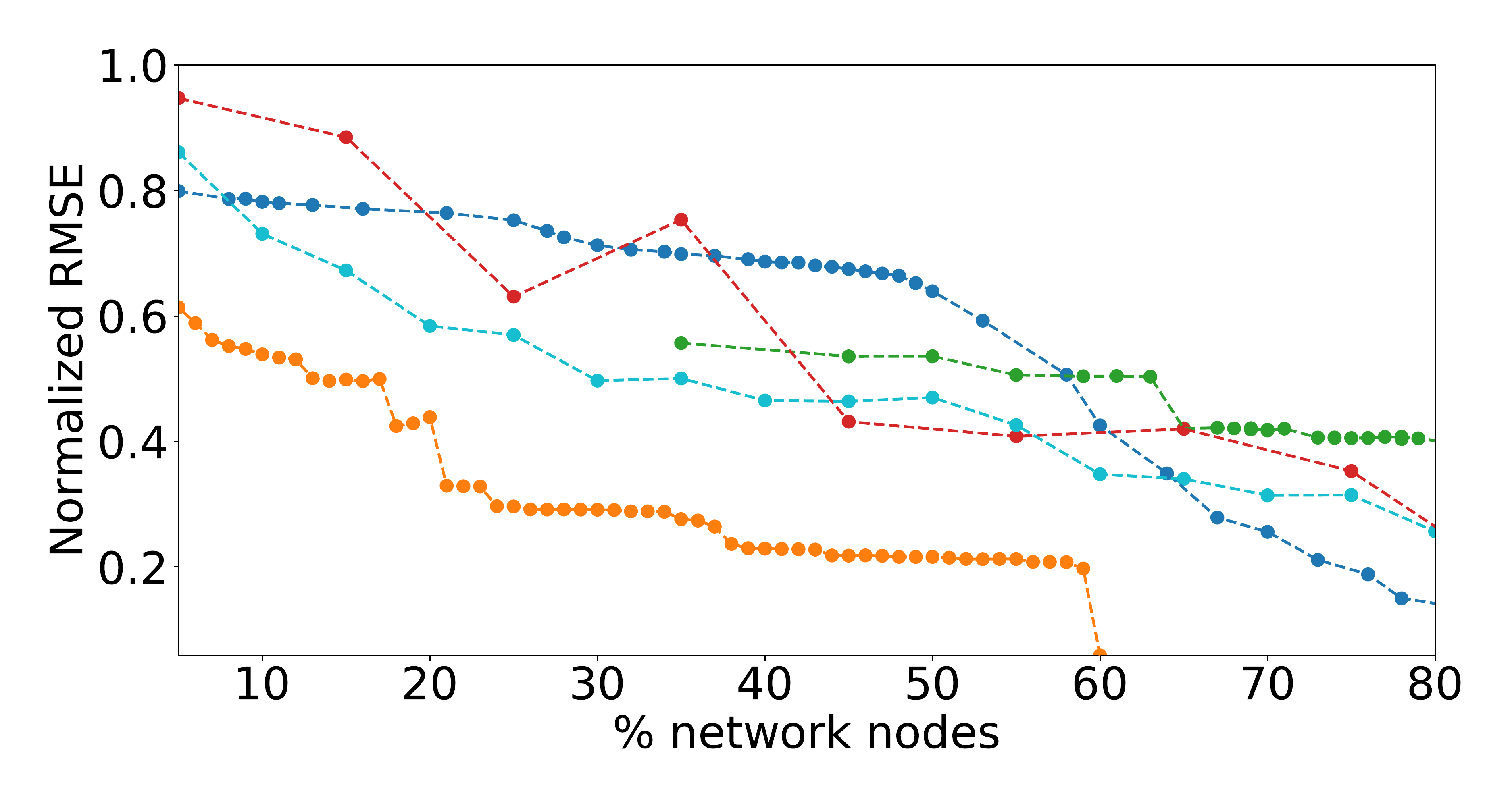}
      Worst case scenario.
    \end{minipage}
    \hfill
    \begin{minipage}[b]{\columnwidth}
      \centering
      \includegraphics[width=\textwidth]{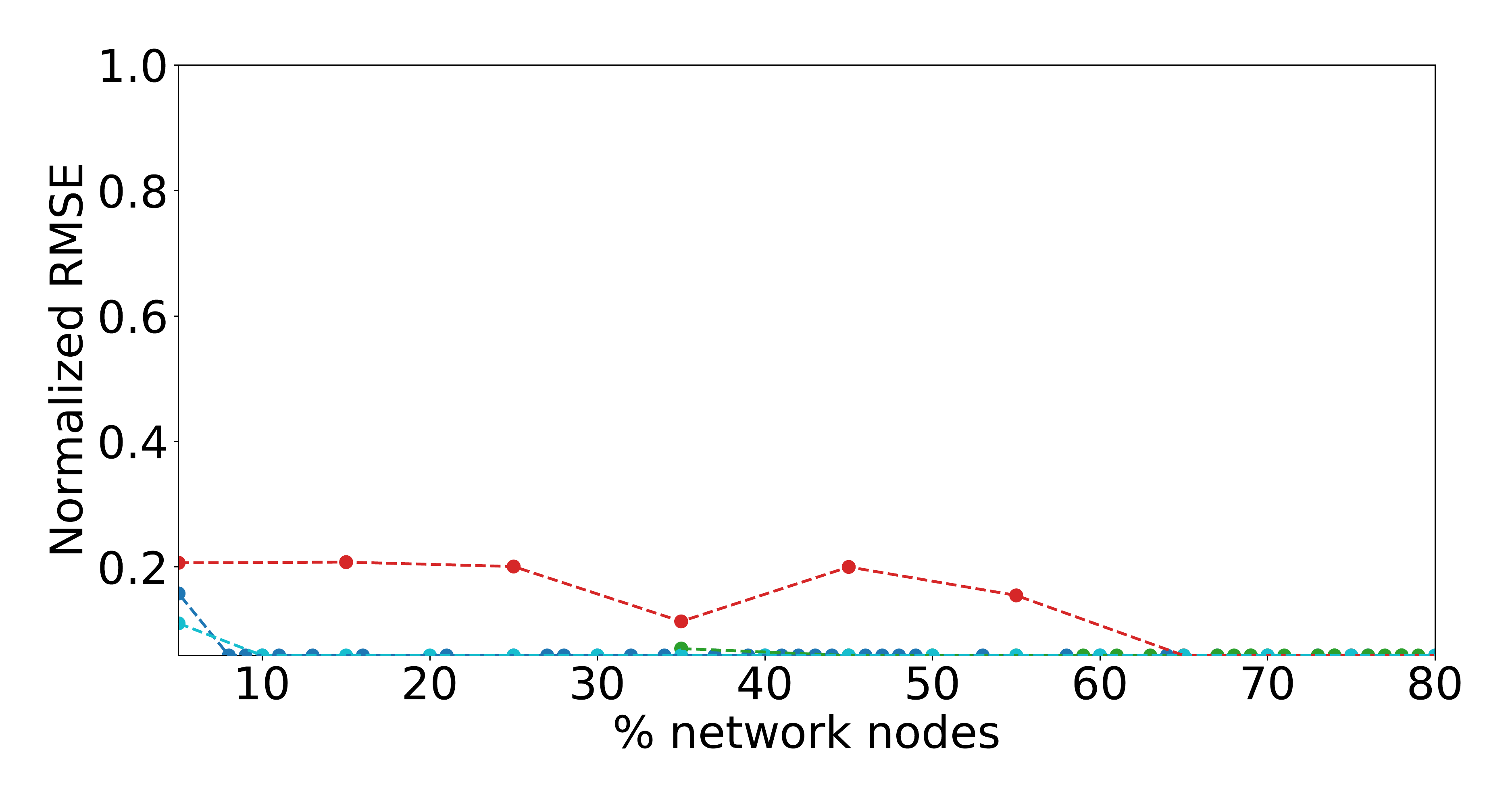}
      Best case scenario.
    \end{minipage}
    \hfill
    \caption{Normalised RMSE of the reconstructed dynamics using different sampling approaches and different dataset size.}
    \label{fig:rmse_dynamics_examples}
\end{figure}

In Fig. \ref{fig:rmse_dynamics_examples} is shown the error (normalised RMSE) of the reconstructed dynamics in 3 junctions, using datasets of different size and different sampling techniques. Intuitively, the higher is the acceptable error in the dynamics reconstruction and the lower is the required number of nodes to be monitored.


\subsection{Sensitivity and Specificity of the General Approaches}
\label{sec:sensitivity_and_specificity}
\textbf{Sensitivity} (also called true positive rate or probability of detection) measures the percentage of junctions with correctly reconstructed dynamics when there is contamination.
\textbf{Specificity} (also called true negative rate) measures the proportion of not polluted junctions (thus with constant dynamics, equal to 0) that are correctly identified as such (thus, indirectly, the probability of false alarms).

The results of this analysis are shown in Table \ref{tab:true_positives} (sensitivity) and Table \ref{tab:true_negatives} (specificity). This study confirms that the GFT frequent nodes dataset (GFT-F) outperforms the other sampling techniques. While the specificity is always high (probability of false alarms is very low), the sensitivity varies significantly. In particular, if 75\% of the WDN nodes is monitored, 51\% of the dynamics are correctly reconstructed with high accuracy. The percentage of correctness reaches 88\% with medium accuracy and 98\% with low accuracy. When 50\% of the network is monitored, we can correctly reconstruct the dynamics in 47\% of the nodes with high accuracy. The percentage increases if medium accuracy (85\%) or low accuracy (90\%) are acceptable. When a low accuracy reconstruction of the signal is sufficient, we obtain good percentage of correct dynamics reconstructed even with 30\% of monitored nodes (70\% of dynamics correctly reconstructed) and 10\% of the WDN nodes (36\% of dynamics correctly reconstructed).

It is worth noting that, while some of the percentages of dynamics correctly reconstructed may appear low, the goal of the WDN monitoring is to guard the most sensible parts of the network (e.g., the most sensible to attacks or accidental contamination). For this reason, even the solution that monitors 36\% of the network installing sensors in 10\% of it may be extremely useful, especially when keeping the sensor installation and maintenance cost low is a priority. Reconstruct the dynamics in central nodes may be sufficient to understand when there is a chemical contamination and where it is spreading. This permits to act consequently and take the required precautions (e.g., closing gates to protect part of the network, alert population). Moreover, if significantly important junctions are known (e.g., junctions close to highly populated areas, chemical plants that may accidentally discharge contamination, etc.), the described approaches can be designed to reconstruct the dynamics with high accuracy in these junctions. Another possible solution could be an hybrid approach that uses more nodes belonging to GFT datasets of potentially dangerous nodes, or that prioritize the dynamics reconstruction of sensible nodes.

\begin{table}[!ht]
\centering
\caption*{High accuracy reconstruction}
\begin{tabular}{ |c|c|c|c|c| }
 \hline
   & 75\% WDN & 50\% WDN & 30\% WDN & 10\% WDN \\
 \hline
 \hline
 GFT-F & 51\% & 47\% & 44\% & 19\% \\ 
 GFT-I & 40\% & 38\% & 15\% & N/A  \\ 
 Lapl. & 32\% & 19\% & 18\% & 10\% \\
 \hline
\end{tabular}
\medskip
\caption*{Medium accuracy reconstruction}
\begin{tabular}{ |c|c|c|c|c| }
 \hline
   & 75\% WDN & 50\% WDN & 30\% WDN & 10\% WDN \\
 \hline
 \hline
 GFT-F & 88\% & 85\% & 63\% & 26\% \\ 
 GFT-I & 81\% & 55\% & 22\% & N/A  \\ 
 Lapl. & 72\% & 41\% & 36 & 15\% \\
 \hline
\end{tabular}
\medskip
\caption*{Low accuracy reconstruction}
\begin{tabular}{ |c|c|c|c|c| }
 \hline
   & 75\% WDN & 50\% WDN & 30\% WDN & 10\% WDN \\
 \hline
 \hline
 GFT-F & 98\% & 90\% & 70\% & 36\% \\ 
 GFT-I & 92\% & 72\% & 43\% & N/A  \\ 
 Lapl. & 91\% & 72\% & 68\% & 37\% \\
 \hline
\end{tabular}

\caption{Sensitivity (true positives): percentage of polluted junctions with correctly reconstructed dynamics using different sensor placement techniques: GFT frequent nodes dataset (GFT-F), GFT important nodes dataset (GFT-I) and Laplacian ranking. 3 different levels of accuracy are used.}
\label{tab:true_positives}
\end{table}

\begin{table}[!ht]
\centering
\caption*{High accuracy reconstruction}
\begin{tabular}{ |c|c|c|c|c| }
 \hline
   & 75\% dataset & 50\% dataset & 30\% dataset & 10\% dataset \\
 \hline
 \hline
 Freq. nodes & 94\% & 94\% & 95\% & 65\% \\ 
 Imp. nodes & 90\% & 92\% & 79\% & N/A  \\ 
 Lap. nodes & 91\% & 91\% & 92\% & 92\% \\
 \hline
\end{tabular}
\medskip
\caption*{Medium accuracy reconstruction}
\begin{tabular}{ |c|c|c|c|c| }
 \hline
   & 75\% dataset & 50\% dataset & 30\% dataset & 10\% dataset \\
 \hline
 \hline
 Freq. nodes & 99\% & 98\% & 97\% & 81\% \\ 
 Imp. nodes & 99\% & 97\% & 88\% & N/A  \\ 
 Lapl. nodes & 99\% & 97\% & 96\% & 96\% \\
 \hline
\end{tabular}
\medskip
\caption*{Low accuracy reconstruction}
\begin{tabular}{ |c|c|c|c|c| }
 \hline
   & 75\% dataset & 50\% dataset & 30\% dataset & 10\% dataset \\
 \hline
 \hline
 Freq. nodes & 100\% & 99\% & 99\% & 99\% \\ 
 Imp. nodes & 99\% & 98\% & 98\% & N/A  \\ 
 Lap. nodes & 99\% & 99\% & 98\% & 98\% \\
 \hline
\end{tabular}

\caption{Specificity (true negatives): percentage of not polluted junctions correctly identified using different  sensor placement techniques: GFT frequent nodes dataset (GFT-F), GFT important nodes dataset (GFT-I) and Laplacian ranking. 3 different levels of accuracy are used.}
\label{tab:true_negatives}
\end{table}

\section{Conclusion}
In this work, we proposed an innovative methodology to reconstruct the dynamics of the chemical diffusion in the WDNs, optimising the number of sensor required. This methodology outperforms the current state of the art both in terms of number of sensors and reconstruction accuracy. The dynamics are reconstructed using graph Fourier transform driven neural networks.

Two main approaches have been analysed: known and unknown source of possible contamination. On one hand, when the possible source is known (e.g., monitoring industrial areas), we optimise the sensor placement identifying the optimal nodes through GFT analysis and we further reduce the number of monitored points using neural networks. With this approach, we are able to reconstruct the dynamics with high accuracy using 20\% of the GFT dataset. In other words, sensors have to be installed in around 5-10\% of all the WDN nodes, depending on the injection point and the network structure. On the other hand, when the possible contamination source is unknown (e.g., general water contamination monitoring), several GFT datasets are used to identify an initial ensemble of important nodes. Also in this case, the number of monitored points is further reduced using a neural networks approach. By monitoring 75\% of the WDN nodes, we are able to reconstruct the dynamics with a specificity up to 98\% (low accuracy reconstruction) and up to 51\% (high accuracy reconstruction). In applications where a low accuracy reconstruction of the signal is sufficient, we obtain good sensitivity even with 30\% of WDN nodes monitored (70\% of dynamics correctly reconstructed) and 10\% of the WDN nodes monitored (36\% of dynamics correctly reconstructed).

As already stated, in many real applications, reconstruct the dynamics in core parts of the WDNs may be sufficient to understand when and where there is a contamination and where it is spreading. This allows to act consequently (e.g., closing gates to protect part of the network, alert population). For this reason, an approach that provides low accuracy but with much less sensors to be installed and monitored could be the best trade-off between cost and monitoring performance.

The proposed sampling techniques are useful beyond the application of WDNs and they can be applied to a variety of infrastructure sensing. They are also useful in the context of digital twin modeling. Future work will focus on how to improve the prediction accuracy using different machine learning techniques and how to further reduce the number of sensors, for example optimising the reconstruction of the dynamics only in the most important and sensible areas of the networks.

\section*{Contributions}
A.P. developed the neural network and conducted the water distribution network simulation. R.S. advised on the neural network research. Z.K. developed the graph Fourier transform. A.P., Z.K. and W.G. wrote the paper.

\section*{Acknowledgements}
The authors (A.P., R.S. \& W.G.) acknowledge funding from the Lloyd's Register Foundation's Programme for Data-Centric Engineering at The Alan Turing Institute. The authors (A.P., R.S. \& W.G.) acknowledge funding from The Alan Turing Institute under the EPSRC grant EP/N510129/1. The author (W.G.) acknowledge funding from EPSRC grant EP/R041725/1. \\
The authors acknowledge Microsoft Corporation for providing cloud resources on Microsoft Azure.

\ifCLASSOPTIONcaptionsoff
  \newpage
\fi

\bibliographystyle{IEEEtran}






\end{document}